\documentclass{article}

\usepackage{microtype}
\usepackage{graphicx}
\usepackage{subfigure}
\usepackage{booktabs} 

\usepackage{hyperref}
\usepackage{xurl} 
\hypersetup{
    breaklinks=true,
    hidelinks
}
\usepackage{siunitx}
\usepackage{amsmath}
\usepackage{longtable}
\usepackage{array}
\usepackage{graphicx} 
\usepackage{twemojis} 
\usepackage{adjustbox} 
\usepackage{pdflscape} 
\usepackage{enumitem}
\usepackage{float} 

\usepackage[accepted]{icml2025}

\usepackage{amsmath}
\usepackage{amssymb}
\usepackage{mathtools}
\usepackage{amsthm}

\usepackage[capitalize,noabbrev]{cleveref}

\theoremstyle{plain}

\theoremstyle{definition}

\theoremstyle{remark}

\usepackage[textsize=tiny]{todonotes}

\newcommand{\llamathree}{\adjustbox{valign=c}{\includegraphics[height=0.9em]{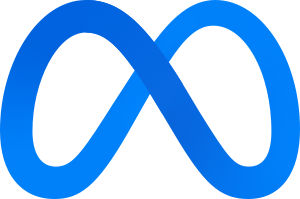}}}
\newcommand{\deepseekvthree}{\adjustbox{valign=c}{\includegraphics[height=1em]{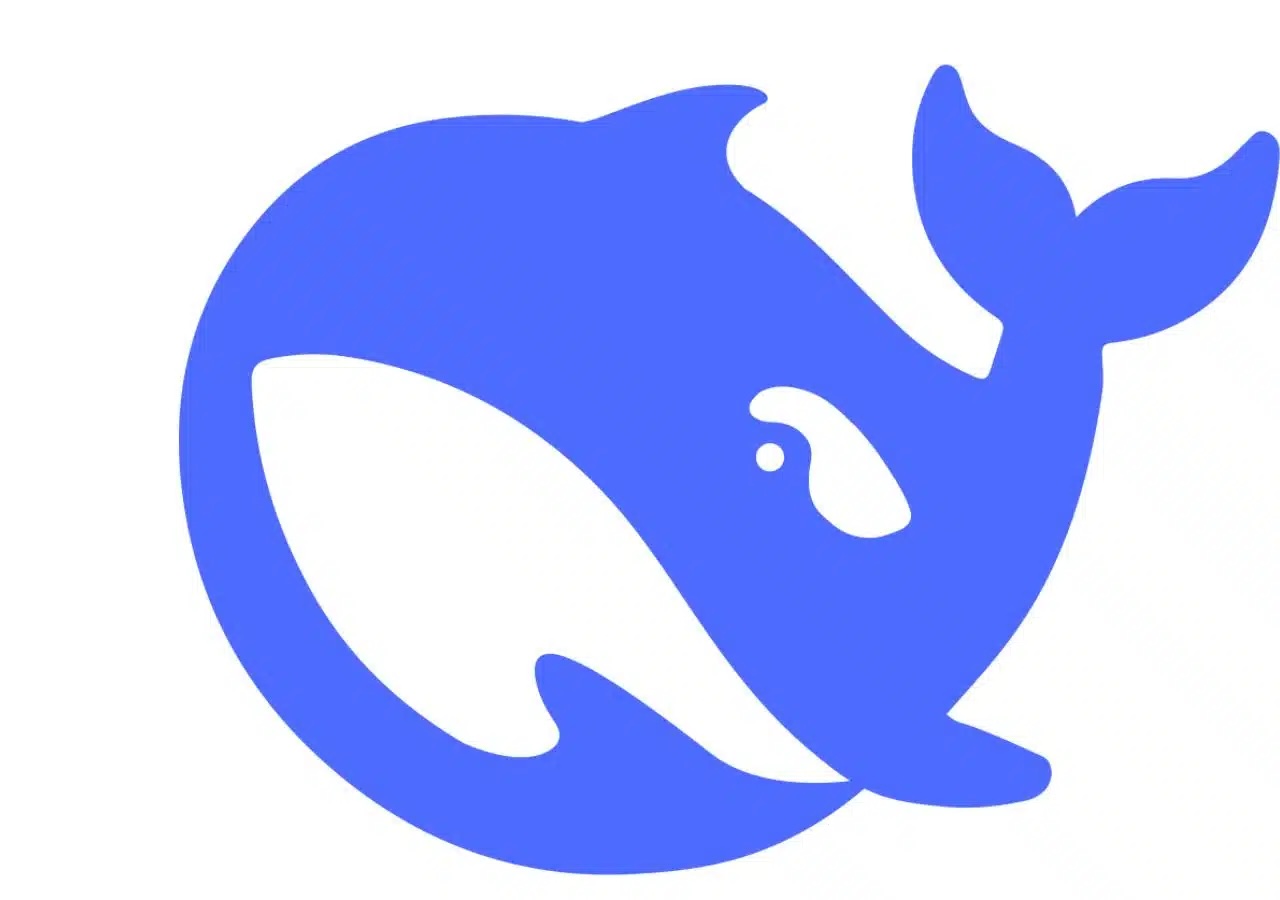}}}

\newcommand{\negligible}{\multicolumn{1}{c}{Negligible}}

\newcommand{\nodata}{\multicolumn{1}{c}{---}} 

\newcommand{\mathequivcheck}{$\star$} 

\icmltitlerunning{Compute Requirements for Algorithmic Innovation in Frontier AI Models}

\begin{document}

\twocolumn[
\icmltitle{Compute Requirements for Algorithmic Innovation in Frontier AI Models}



\icmlsetsymbol{equal}{*}

\begin{icmlauthorlist}
\icmlauthor{Peter Barnett}{comp}
\end{icmlauthorlist}

\icmlaffiliation{comp}{Machine Intelligence Research Institute, CA, USA} 

\icmlcorrespondingauthor{Peter Barnett}{peter@intelligence.org} 

\icmlkeywords{Machine Learning, ICML}

\vskip 0.3in
]



\printAffiliationsAndNotice{}  

\begin{abstract}
Algorithmic innovation in the pretraining of large language models has driven a massive reduction in the total compute required to reach a given level of capability. In this paper we empirically investigate the compute requirements for developing algorithmic innovations. We catalog 36 pre-training algorithmic innovations used in Llama 3 and DeepSeek-V3. For each innovation we estimate both the total FLOP used in development and the FLOP/s of the hardware utilized. Innovations using significant resources double in their requirements each year. We then use this dataset to investigate the effect of compute caps on innovation. 
Our analysis suggests that compute caps alone are unlikely to dramatically slow AI algorithmic progress. Even stringent compute caps---such as capping total operations to the compute used to train GPT-2 or capping hardware capacity to 8 H100 GPUs---could still have allowed for half of the cataloged innovations. 
\end{abstract}

\section{Introduction}
The control of computing resources is a central lever in governing the development of AI~\citep{sastry2024computing}. This includes setting training compute thresholds above which training must be reported, monitored, and potentially banned~\cite{heim2024training,miotti_narrow_2024,aguirre2025keep}. Nations may also use control over compute to limit their rivals' AI progress, or to prevent malicious non-state actors from gaining access to dual-use AI systems~\cite{heim2024governing, scher_mechanisms_2024}. 

AI capabilities can be increased by spending more compute~\cite{hoffmann_training_2022} (training larger models on more data) and by using more efficient algorithms. The development of increasingly efficient algorithms is referred to as \emph{algorithmic progress} and is the focus of this paper. 

Continued algorithmic progress may limit the effectiveness of compute governance. If less compute is needed to develop dangerous AI capabilities, it may not be possible for nations to monitor all relevant compute. Currently, the amount of compute needed to reach a given level of capability declines by a factor of approximately 3 each year~\citep{ho2024algorithmic}. 

However, algorithmic progress itself also depends on access to compute; researchers must run experiments to develop and validate algorithmic innovations. Hence restricting compute may slow algorithmic progress. 

The term ``compute" may be ambiguous; in this paper we discuss both:
\begin{itemize}[nosep]
    \item \emph{Total operations}: The number of operations used, measured in FLOP.
    \item \emph{Hardware capacity}: The number of operations per second on the available hardware, measured in TFLOP/s and determined by the number and type of accelerators (e.g., GPUs, TPUs) used. 
\end{itemize}
Some algorithmic innovations require many total operations to develop, such as when the development requires training large models. Other innovations can require very few total operations, but do require a large hardware capacity. For example, parallelization improvements may not require whole models to be trained, but do require many GPUs to test; these innovations then allow hardware to be used more effectively when training models. We consider a diverse range of innovations beyond merely improving training loss efficiency; we include improvements in hardware utilization, model context length, and various other areas.

\begin{table}[h!] 
\centering
\caption{Examples of cataloged algorithmic innovations and their estimated total operations (FLOP) and hardware capacity (TFLOP/s). Full list in \cref{app:table_of_innovations}.}
\label{tab:intro_examples}
\vskip 0.15in 

\setlength{\tabcolsep}{3pt} 

\begin{small} 
\begin{tabular}{@{}p{5.5cm} S[table-format=1.2e1] S[table-format=5, table-auto-round]@{}} 
\toprule
\bfseries Innovation & {\bfseries FLOP} & {\bfseries TFLOP/s} \\
\midrule
Byte-pair encoding \citep{sennrich_neural_2016} & 9.00e18 & 22.6 \\ 
Transformer \citep{vaswani_attention_2017}       & 4.00e19 & 84.8 \\
AdamW \citep{loshchilov_decoupled_2018}          & 1.76e19 & 329.1 \\
RMSNorm \citep{zhang_root_2019}                 & 1.37e20 & 149.4 \\ 
ZeRO \citep{rajbhandari_zero_2020}              & \multicolumn{1}{c}{Negligible} & 50000 \\ 
FlashAttention \citep{dao_flashattention_2022}   & 6.22e20 & 2540 \\ 
\bottomrule
\end{tabular}
\end{small}
\end{table}
\vspace{0.6cm}
In this paper, we undertake the first empirical investigation of the relationship between compute and the development of algorithmic innovations used in open frontier models. Examples of these innovations are shown in \cref{tab:intro_examples}. We then investigate the impact of compute caps (restrictions on compute) by looking at how many of these innovations could have been developed under various restrictions. 

\section{Methodology}
We catalog pre-training algorithmic innovations implemented in two prominent open models: Llama 3 \cite{grattafiori_llama_2024} and DeepSeek-V3~\cite{deepseek-ai_deepseek-v3_2025}. We focus on innovations which were actually implemented, as opposed to highly cited. We include innovations that ``inspired" modifications if these are cited. 

Llama 3 is based on Llama 2~\cite{touvron2023llama2} which in turn was based on LLaMA 1~\cite{touvron2023llama1}, DeepSeek-V3 is largely based on DeepSeek-V2~\cite{deepseek-ai_deepseek-v2_2024}, and so we focus on the algorithmic innovations implemented in each of these five models. These models were chosen because they were state-of-the-art for open model capabilities when released, and the developers have provided extensive details about their training. 

For each innovation, we review the original paper, and calculate the total operations used for the experiments in the paper, and the total terraFLOP/s (TFLOP/s) of the hardware used (the hardware capacity). Calculating total operations primarily involves calculating the compute used to train each model trained in the paper, and then summing these values. 
Some compute values were obtained from personal correspondence with paper authors. 
In various innovations, the compute used was negligible, or the original paper did not provide sufficient information to determine the FLOP or TFLOP/s. 

For example, to estimate compute for the original Transformer architecture~\citep{vaswani_attention_2017}, we calculate and sum the total operations used to train all models trained as part of the paper, including ablation studies. The paper also states that all models were trained on one machine with 8 NVIDIA P100 GPUs.

We additionally classify the innovations into categories: 
\begin{itemize}[nosep]
    \item Architecture: Modifications to model architecture (e.g., layers, activations, position embeddings).
    \item Data \& Tokenization: Methods for data processing, tokenization, or selection.
    \item Efficiencies: Techniques reducing cost (FLOP, memory, time) per step (e.g., low precision, faster algorithms).
    \item Scaling laws: Experiments to determine how to allocate compute while scaling models.
    \item Scaling training: Techniques for efficient large-scale distributed training (e.g., parallelism strategies).
    \item Training: Changes to the training process (e.g., optimizers, loss functions). 
\end{itemize}
\subsection{Negligible FLOP Innovations}
Various innovations are more efficient implementations of mathematically equivalent operations~\cite{rabe_self-attention_2022, korthikanti_reducing_2023, dao_flashattention_2022}. These innovations do not require large training runs to validate, because the result would be mathematically identical, but do require significant hardware capacity to be tested at scale (e.g., with a relatively small number of forward and backward passes). 9 out of 36 (25\%) of the innovations cataloged used negligible FLOP. 

\section{Trends}
\begin{figure}[h]
    \centering
    \includegraphics[width=1.0\linewidth]{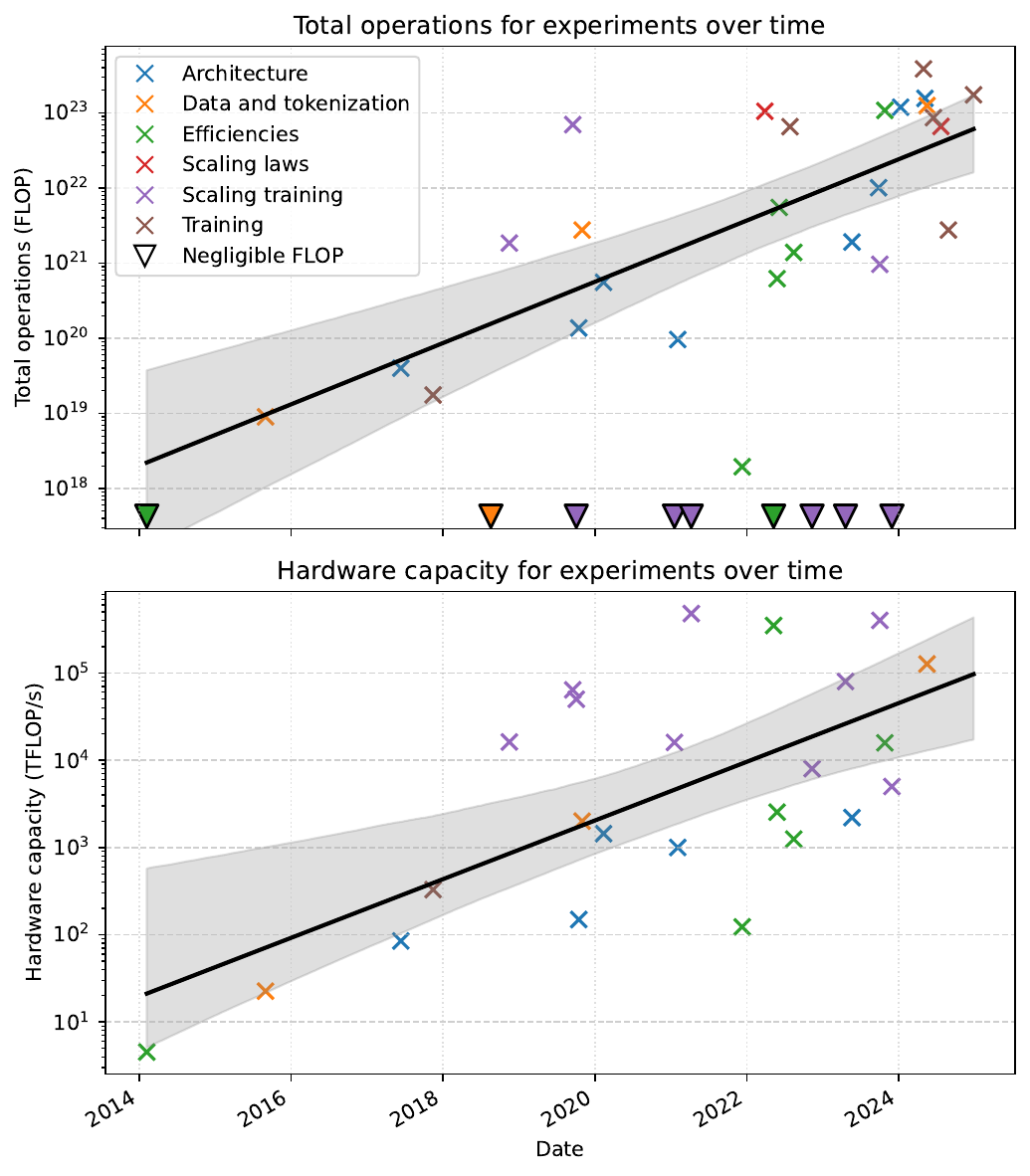}
    \caption{Top: The total operations (in FLOP) for the different algorithmic innovations over time, colored by their category.  The triangles indicate innovations which used negligible FLOP in their experiments. The black line shows the trend, and the shaded area indicates the 95\% CI for this trend. This trend line is only for innovations that did not use negligible compute.\\
    Bottom: The hardware capacity (in TFLOP/s) for the different algorithmic innovations. The black line shows the trend, and the shaded area indicates the 95\% CI for this trend.}
    \label{fig:compute_trends}
\end{figure}

We first investigate how the compute requirements for algorithmic innovations are changing over time (\cref{fig:compute_trends}). For experiments which do not use negligible compute, the FLOP used is increasing at a rate of $\times$2.53/year (95\% CI: 1.86--3.38). The hardware capacity in TFLOP/s is increasing at a rate of $\times$2.14/year (95\% CI: 1.44--2.76). Pre-training compute performance per dollar (measured in TFLOP/s per dollar) is increasing at a rate of around $\times$1.3/year~\cite{EpochMachineLearningHardware2024}. Both the total operations for innovations and the hardware capacity are increasing faster than this, implying that the increased compute use is not due purely to cheaper compute. 

\begin{figure}[h]
    \centering
    \includegraphics[width=1.0\linewidth]{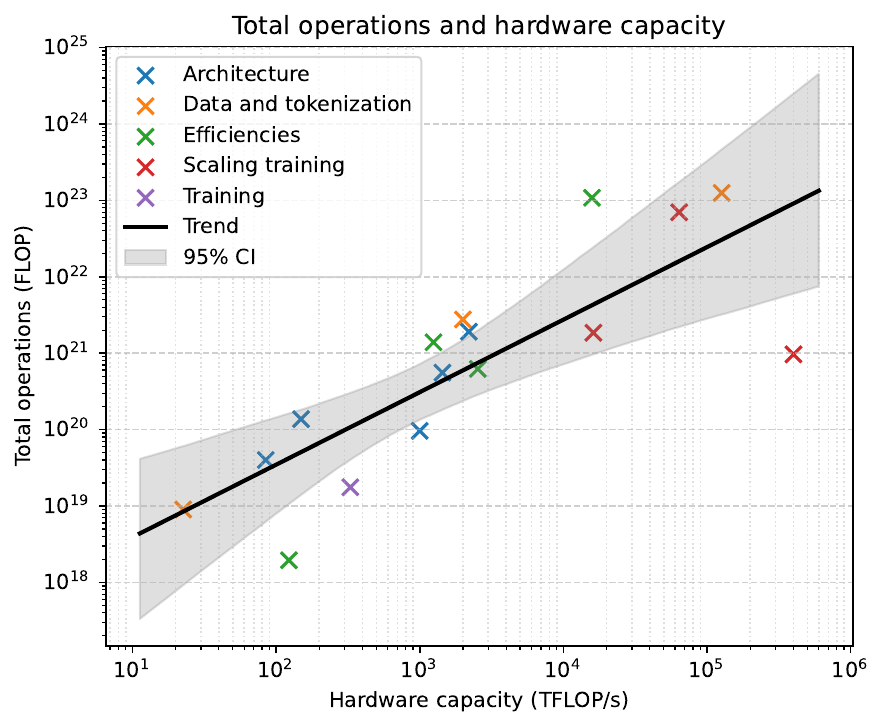}
    \caption{Total operations used for experiments versus hardware capacity for the different algorithmic innovations. Compute used is correlated with hardware capacity. This graph only includes innovations which used non-negligible FLOP. }
    \label{fig:flop_vs_tflops}
\end{figure}

The increase in FLOP used for experiments is likely at least partially due to increased hardware capacity. \cref{fig:flop_vs_tflops} shows that the total operations used in the experiments for innovations is correlated with  
hardware capacity.  

\section{Impact of Compute Caps}
Compute caps may reduce the rate of algorithmic progress. Along with the direct impact of preventing the development of dangerous AI systems, caps may also increase the time it takes until dangerous AI systems can be developed with sub-threshold levels of compute. We consider two implementations of compute caps: restrictions on the total operations (\emph{FLOP caps}), and restrictions on the hardware capacity (\emph{hardware caps}). 

FLOP caps would entail capping the total number of computational operations an actor (such as a researcher or AI developer) can use in a period of time. The budget could be reset at regular intervals (e.g., allowing $10^{21}$ FLOP per month), or after audits confirmed that operations were not being used for prohibited activities. This could work by allocating centralized cloud compute~\citep{heim2024governing}, or having on-chip mechanisms to count the number of operations and enforce usage limits~\cite{WR-A3056-1, petrie2024interim}.

Alternatively, hardware caps could involve restricting the hardware that actors are allowed to own and use. This could be measured by the combined TFLOP/s of their AI-specific hardware. For example, a hardware cap of 5000 TFLOP/s would allow actors to own and use 5 H100 GPUs or 16 A100 GPUs. Hardware caps could be enabled by implementing location verification mechanisms on AI chips combined with a centralized chip registry~\cite{brass2024location}. 

\begin{figure}[h]
    \centering
    \includegraphics[width=1\linewidth]{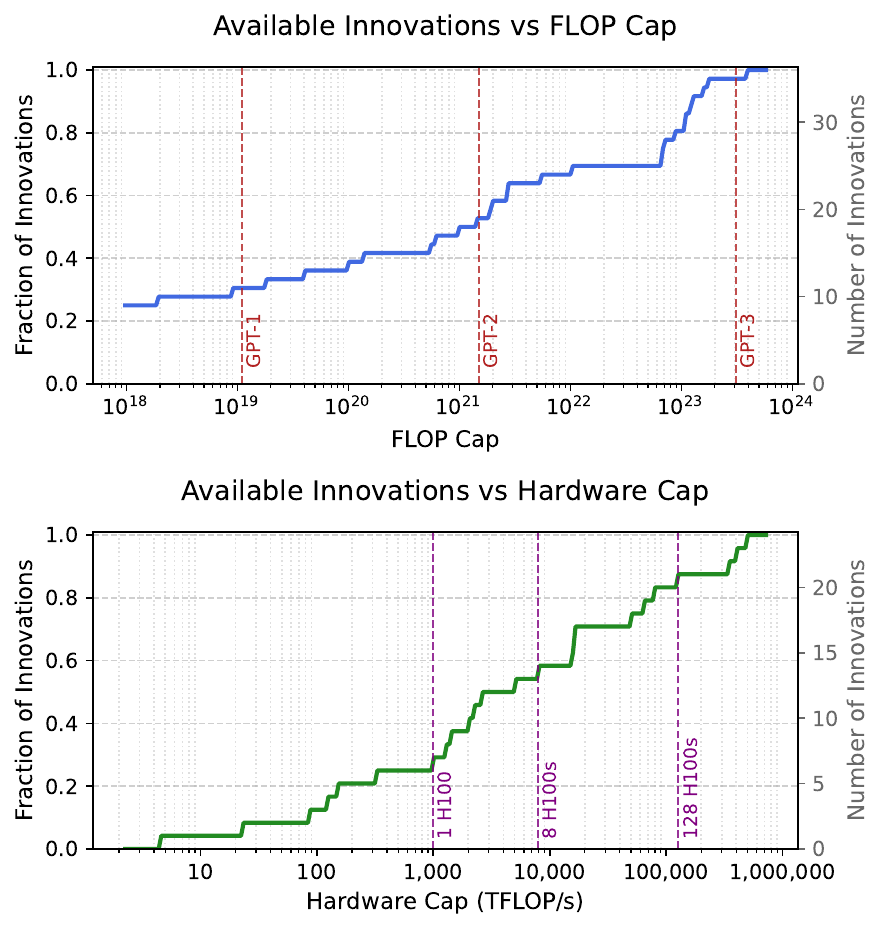}
    \caption{Top: The fraction (left axis) and cumulative count (right axis) of innovations whose total operations falls below a given cap value. Each point on the line answers: “If regulators limited training runs to $\leq X$ FLOP, how many of the techniques in this dataset would still have been discoverable?” For reference, we show the training compute for GPT-1, 2 and 3~\citep{EpochNotableModels2024}. \\
    Bottom: Identical analysis for hardware caps. For reference, we show the hardware capacity of different numbers of H100 GPUs. \\
    Moving left-to-right shows how the pool of accessible innovations grows as caps loosen. 
    }
    \label{fig:compute_cap_effects}
\end{figure}

As a preliminary measure of the impact of compute caps on algorithmic progress, we calculate what fraction of the cataloged innovations fall below a given FLOP or hardware cap (\cref{fig:compute_cap_effects}). That is: what fraction of the innovations could have been developed if a compute cap had been enforced at a certain level? This estimate will overstate the effects of compute caps, as researchers could attempt to be more efficient with the total operations that they are allocated or run their hardware for longer. However, the x-axes of these plots are on a log-scale, and so even doubling the effective compute cap would have a relatively small effect.

The results in \cref{fig:compute_cap_effects} show that \textbf{compute caps alone are unlikely to dramatically slow AI algorithmic progress}. Even very restrictive caps such as capping total operations to the compute used to train GPT-2 or capping hardware capacity to 8 H100s would both only disallow about half of the cataloged innovations. 

A further implication is that US-led export controls on AI hardware are unlikely to slow rates of AI algorithmic progress in rival states. The existing hardware in these states is likely sufficient for current algorithmic innovation. 

As observed in \cref{fig:compute_trends}, the total operations and hardware capacity are growing over time; if these trends continue to 2028, the median hardware capacity would be $10^{6}$~TFLOP/s (around 1000 H100 GPUs) and the median total operations for innovations using non-negligible FLOP would be $10^{24}$~FLOP. Compute caps at these levels may therefore slow AI algorithmic progress. These caps are would still be fairly low, as they are 10--100$\times$ lower than the hardware and total operations used to train frontier AI models in 2025. 

\section{Limitations and Future Work}
\label{sec:limitations_future_work}
While this study provides initial estimates for the compute requirements of algorithmic innovations, there are several  limitations inherent in our methodology and scope. We also propose avenues for future research to build upon this work.
\subsection{Limitations}

\textbf{Focus on Reported Successes:} Our analysis relies on the experiments described in the final publications introducing an innovation. This approach misses the potentially large amount of compute spent on unsuccessful experiments and preliminary explorations that did not make it into the published results. The compute costs reported in this paper are a \emph{lower bound} on the actual compute used. Although, because researchers could likely be more efficient with their compute, these are not lower bounds on the compute \emph{required} to develop these innovations.

\textbf{Focus on Initial Research:} We focus on estimating compute used in the initial research introducing an innovation. However, validating innovations at scale may often require much more compute than is used initially, and only some cataloged papers include large-scale validation. Our estimates may therefore miss follow-up work (often spread across multiple papers) that demonstrates an innovation continues to work with scale. 

\textbf{Ignoring Research Lineage:} We estimate the compute associated with the specific paper introducing or validating an innovation. This does not account for the cumulative compute of the prior work upon which that innovation builds.

\textbf{Exclusion of Proprietary Models:} Our reliance on open models means we miss innovations developed and implemented in closed, proprietary systems (e.g., at companies like Google DeepMind, OpenAI, Anthropic). These labs often operate at the largest scales, and their internal innovations could have different compute requirements.

\textbf{Pre-training Focus:} We deliberately restricted our scope to pre-training innovations. Substantial algorithmic progress also occurs in post-training techniques (e.g., instruction fine-tuning~\citep{ouyang2022training}, RLHF~\citep{stiennon2020learning}, reinforcement learning for reasoning capabilities~\citep{jaech2024openai, guo2025deepseek}), whose compute requirements are not analyzed in this paper. Currently, post-training generally uses much less compute than pre-training~\citep{davidson2023ai}, so it may also be that experiments to develop algorithmic innovations for pre-training require significantly more compute than for post-training innovations. However, this may not continue to hold as increasing compute is spent on reinforcement learning.

\textbf{Data Availability and Reporting Bias:} The estimation of total operations and hardware capacity relies  on the details provided in the source papers. This information is not always available or reported consistently, leading to missing data points. There may be a selection effect where certain researchers are more likely or able to report the hardware used, potentially biasing our view of typical requirements. This includes research from industry developers, who may not want to disclose details about their companies' hardware capabilities. Our estimates are approximations based on available data.

\textbf{Citation Bias:} Developers are more likely to cite work and innovations from their own organization. For example, 5 out of 17 innovations cataloged for DeepSeek-V3 were developed by DeepSeek. Because of this, the proportion of Chinese innovations in \cref{app:table_of_innovations} is likely overstated.

\textbf{Compute Abundance and Scarcity:} This analysis examines past trends under conditions of relative compute abundance for leading labs. It does not model how research might adapt if compute were restricted. Under this regime, researchers would be incentivized to use their compute more efficiently. Under conditions of relative scarcity, researchers would likely run experiments at smaller scales and halt experiments earlier if they did not show promising results. This would likely also change the type of innovations being developed in order to respond to the restrictions (``scarcity breeds creativity"). It is plausible that strong compute caps could incentivize a shift towards innovations that require less compute to develop, and are more useful for AI systems which require less compute to train.

\subsection{Future Work}
This paper demonstrates the feasibility and potential utility of analyzing the compute requirements for algorithmic innovation. This is a tractable and important research direction for understanding the dynamics of AI development and informing governance. Future work could expand and refine this analysis in several ways:

\textbf{Qualitative Validation:} Complement the quantitative analysis derived from papers with qualitative methods. Interviewing AI researchers could provide valuable insights into the ratio of reported compute versus the actual R\&D compute (including failed experiments), helping to calibrate the estimates presented here.

\textbf{Broader Model and Technique Coverage:} Extend the analysis to include other open model series, such as the OLMo models~\citep{groeneveld2024olmo, olmo20242olmo}. Crucially, expand the scope beyond pre-training to include post-training innovations, including the post-training innovations used in the models covered in this paper, as well as other models such as DeepSeek-R1~\citep{guo2025deepseek} and T\"ulu 3~\citep{lambert2024tulu}.

\textbf{Focus on Distributed Training:} Investigate the specific compute costs associated with developing novel \emph{distributed training} algorithms (e.g., methods like DiLoCo~\citep{douillard2023diloco} and those used in INTELLECT-1 \citep{jaghouar2024intellect}). Progress in distributed training could significantly challenge compute governance regimes aimed at limiting training above a certain scale, making the R\&D compute for these techniques important to understand.

\textbf{Quantifying Innovation Impact (CEG):} Connect the \emph{cost} of developing an innovation (as estimated here) with how useful that innovation was. Future work could attempt to estimate a form of Compute-Equivalent Gain (CEG)~\citep{davidson2023ai} specifically attributable to different algorithmic innovations, allowing analysis of the ``return on compute investment'' for different types of algorithmic progress.

\section{Impact Statement}
This research focuses on a key challenge for compute governance: algorithmic progress may eventually allow actors to circumvent restrictions on computational resources, and gain access to dangerously capable AI systems. If algorithmic progress continues at its current pace, compute governance strategies could become ineffective. We hope this work encourages more researchers to investigate this relationship and other critical questions in compute governance.

Our analysis of compute caps suggests that only very tight restrictions would dramatically slow algorithmic progress. However, we caution against interpreting these results as evidence that algorithmic progress cannot be managed. Rather, compute caps likely need to be combined with other approaches, such as legal frameworks, to be effective for this aim. 

We recognize that limiting algorithmic innovation involves  tradeoffs, as restrictions would delay both the risks and benefits of AI advancement. Our primary concern is that unchecked algorithmic progress could undermine governance mechanisms designed for responsible AI development. It is likely worth trading some speed in AI progress in exchange for safer and less destabilizing development.

\bibliography{algorithmic_progress}

\begin{thebibliography}{60}
\providecommand{\natexlab}[1]{#1}
\providecommand{\url}[1]{\texttt{#1}}
\expandafter\ifx\csname urlstyle\endcsname\relax
  \providecommand{\doi}[1]{doi: #1}\else
  \providecommand{\doi}{doi: \begingroup \urlstyle{rm}\Url}\fi

\bibitem[Aguirre(2025)]{aguirre2025keep}
Aguirre, A.
\newblock {Keep the Future Human: Why and How We Should Close the Gates to AGI and Superintelligence, and What We Should Build Instead}.
\newblock \emph{arXiv preprint arXiv:2311.09452}, 2025.
\newblock URL \url{https://arxiv.org/abs/2311.09452v4}.

\bibitem[Ainslie et~al.(2023)Ainslie, Lee-Thorp, Jong, Zemlyanskiy, Lebron, and Sanghai]{ainslie_gqa_2023}
Ainslie, J., Lee-Thorp, J., Jong, M.~d., Zemlyanskiy, Y., Lebron, F., and Sanghai, S.
\newblock {GQA}: {Training} {Generalized} {Multi}-{Query} {Transformer} {Models} from {Multi}-{Head} {Checkpoints}.
\newblock December 2023.
\newblock URL \url{https://openreview.net/forum?id=hmOwOZWzYE}.

\bibitem[Bauer et~al.(2014)Bauer, Treichler, and Aiken]{bauer_singe_2014}
Bauer, M., Treichler, S., and Aiken, A.
\newblock Singe: leveraging warp specialization for high performance on {GPUs}.
\newblock \emph{SIGPLAN Not.}, 49\penalty0 (8):\penalty0 119--130, February 2014.
\newblock ISSN 0362-1340.
\newblock \doi{10.1145/2692916.2555258}.
\newblock URL \url{https://doi.org/10.1145/2692916.2555258}.

\bibitem[Bavarian et~al.(2022)Bavarian, Jun, Tezak, Schulman, McLeavey, Tworek, and Chen]{bavarian_efficient_2022}
Bavarian, M., Jun, H., Tezak, N., Schulman, J., McLeavey, C., Tworek, J., and Chen, M.
\newblock Efficient {Training} of {Language} {Models} to {Fill} in the {Middle}, July 2022.
\newblock URL \url{http://arxiv.org/abs/2207.14255}.
\newblock arXiv:2207.14255 [cs].

\bibitem[Brass \& Aarne(2024)Brass and Aarne]{brass2024location}
Brass, A. and Aarne, O.
\newblock Location verification for ai chips.
\newblock Issue brief, Institute for AI Policy and Strategy (IAPS), April 2024.
\newblock URL \url{https://www.iaps.ai/research/location-verification-for-ai-chips}.

\bibitem[Dai et~al.(2024)Dai, Deng, Zhao, Xu, Gao, Chen, Li, Zeng, Yu, Wu, et~al.]{dai2024deepseekmoe}
Dai, D., Deng, C., Zhao, C., Xu, R., Gao, H., Chen, D., Li, J., Zeng, W., Yu, X., Wu, Y., et~al.
\newblock Deepseekmoe: Towards ultimate expert specialization in mixture-of-experts language models.
\newblock \emph{arXiv preprint arXiv:2401.06066}, 2024.

\bibitem[Dao et~al.(2022)Dao, Fu, Ermon, Rudra, and Ré]{dao_flashattention_2022}
Dao, T., Fu, D.~Y., Ermon, S., Rudra, A., and Ré, C.
\newblock {FLASHATTENTION}: fast and memory-efficient exact attention with {IO}-awareness.
\newblock In \emph{Proceedings of the 36th {International} {Conference} on {Neural} {Information} {Processing} {Systems}}, {NIPS} '22, pp.\  16344--16359, Red Hook, NY, USA, November 2022. Curran Associates Inc.
\newblock ISBN 978-1-71387-108-8.

\bibitem[Davidson et~al.(2023)Davidson, Denain, Villalobos, and Bas]{davidson2023ai}
Davidson, T., Denain, J.-S., Villalobos, P., and Bas, G.
\newblock Ai capabilities can be significantly improved without expensive retraining.
\newblock \emph{arXiv preprint arXiv:2312.07413}, 2023.

\bibitem[DeepSeek-AI et~al.(2024)DeepSeek-AI, Zhu, Guo, Shao, Yang, Wang, Xu, Wu, Li, Gao, Ma, Zeng, Bi, Gu, Xu, Dai, Dong, Zhang, Piao, Gou, Xie, Hao, Wang, Song, Chen, Xie, Guan, You, Liu, Du, Gao, Lu, Chen, Wang, Deng, Li, Zhao, Ruan, Luo, and Liang]{deepseek-ai_deepseek-coder-v2_2024}
DeepSeek-AI, Zhu, Q., Guo, D., Shao, Z., Yang, D., Wang, P., Xu, R., Wu, Y., Li, Y., Gao, H., Ma, S., Zeng, W., Bi, X., Gu, Z., Xu, H., Dai, D., Dong, K., Zhang, L., Piao, Y., Gou, Z., Xie, Z., Hao, Z., Wang, B., Song, J., Chen, D., Xie, X., Guan, K., You, Y., Liu, A., Du, Q., Gao, W., Lu, X., Chen, Q., Wang, Y., Deng, C., Li, J., Zhao, C., Ruan, C., Luo, F., and Liang, W.
\newblock {DeepSeek}-{Coder}-{V2}: {Breaking} the {Barrier} of {Closed}-{Source} {Models} in {Code} {Intelligence}, June 2024.
\newblock URL \url{http://arxiv.org/abs/2406.11931}.
\newblock arXiv:2406.11931 [cs].

\bibitem[Dettmers et~al.(2022)Dettmers, Lewis, Belkada, and Zettlemoyer]{dettmers_llmint8_2022}
Dettmers, T., Lewis, M., Belkada, Y., and Zettlemoyer, L.
\newblock {LLM}.int8(): 8-bit matrix multiplication for transformers at scale.
\newblock In \emph{Proceedings of the 36th {International} {Conference} on {Neural} {Information} {Processing} {Systems}}, {NIPS} '22, pp.\  30318--30332, Red Hook, NY, USA, November 2022. Curran Associates Inc.
\newblock ISBN 978-1-71387-108-8.

\bibitem[Douillard et~al.(2023)Douillard, Feng, Rusu, Chhaparia, Donchev, Kuncoro, Ranzato, Szlam, and Shen]{douillard2023diloco}
Douillard, A., Feng, Q., Rusu, A.~A., Chhaparia, R., Donchev, Y., Kuncoro, A., Ranzato, M., Szlam, A., and Shen, J.
\newblock Diloco: Distributed low-communication training of language models.
\newblock \emph{arXiv preprint arXiv:2311.08105}, 2023.

\bibitem[{Epoch AI}(2024{\natexlab{a}})]{EpochMachineLearningHardware2024}
{Epoch AI}.
\newblock Data on machine learning hardware”, 10 2024{\natexlab{a}}.
\newblock URL \url{https://epoch.ai/data/machine-learning-hardware}.
\newblock Accessed: 2025-04-28.

\bibitem[{Epoch AI}(2024{\natexlab{b}})]{EpochNotableModels2024}
{Epoch AI}.
\newblock Data on notable ai models, 6 2024{\natexlab{b}}.
\newblock URL \url{https://epoch.ai/data/notable-ai-models}.
\newblock Accessed: 2025-05-11.

\bibitem[Gloeckle et~al.(2024)Gloeckle, Idrissi, Rozière, Lopez-Paz, and Synnaeve]{gloeckle_better_2024}
Gloeckle, F., Idrissi, B.~Y., Rozière, B., Lopez-Paz, D., and Synnaeve, G.
\newblock Better \& faster large language models via multi-token prediction.
\newblock In \emph{Proceedings of the 41st {International} {Conference} on {Machine} {Learning}}, volume 235 of \emph{{ICML}'24}, pp.\  15706--15734, Vienna, Austria, July 2024. JMLR.org.

\bibitem[Grattafiori et~al.(2024)Grattafiori, Dubey, Jauhri, Pandey, Kadian, Al-Dahle, Letman, Mathur, Schelten, Vaughan, et~al.]{grattafiori_llama_2024}
Grattafiori, A., Dubey, A., Jauhri, A., Pandey, A., Kadian, A., Al-Dahle, A., Letman, A., Mathur, A., Schelten, A., Vaughan, A., et~al.
\newblock The llama 3 herd of models, 2024.

\bibitem[Groeneveld et~al.(2024)Groeneveld, Beltagy, Walsh, Bhagia, Kinney, Tafjord, Jha, Ivison, Magnusson, Wang, et~al.]{groeneveld2024olmo}
Groeneveld, D., Beltagy, I., Walsh, P., Bhagia, A., Kinney, R., Tafjord, O., Jha, A.~H., Ivison, H., Magnusson, I., Wang, Y., et~al.
\newblock Olmo: Accelerating the science of language models.
\newblock \emph{arXiv preprint arXiv:2402.00838}, 2024.

\bibitem[Guo et~al.(2025)Guo, Yang, Zhang, Song, Zhang, Xu, Zhu, Ma, Wang, Bi, et~al.]{guo2025deepseek}
Guo, D., Yang, D., Zhang, H., Song, J., Zhang, R., Xu, R., Zhu, Q., Ma, S., Wang, P., Bi, X., et~al.
\newblock Deepseek-r1: Incentivizing reasoning capability in llms via reinforcement learning.
\newblock \emph{arXiv preprint arXiv:2501.12948}, 2025.

\bibitem[Heim \& Koessler(2024)Heim and Koessler]{heim2024training}
Heim, L. and Koessler, L.
\newblock Training compute thresholds: Features and functions in ai regulation.
\newblock \emph{arXiv preprint arXiv:2405.10799}, 2024.

\bibitem[Heim et~al.(2024)Heim, Fist, Egan, Huang, Zekany, Trager, Osborne, and Zilberman]{heim2024governing}
Heim, L., Fist, T., Egan, J., Huang, S., Zekany, S., Trager, R., Osborne, M.~A., and Zilberman, N.
\newblock Governing through the cloud: The intermediary role of compute providers in ai regulation.
\newblock \emph{arXiv preprint arXiv:2403.08501}, 2024.

\bibitem[Ho et~al.(2024)Ho, Besiroglu, Erdil, Owen, Rahman, Guo, Atkinson, Thompson, and Sevilla]{ho2024algorithmic}
Ho, A., Besiroglu, T., Erdil, E., Owen, D., Rahman, R., Guo, Z.~C., Atkinson, D., Thompson, N., and Sevilla, J.
\newblock Algorithmic progress in language models, 2024.

\bibitem[Hoffmann et~al.(2022)Hoffmann, Borgeaud, Mensch, Buchatskaya, Cai, Rutherford, Casas, Hendricks, Welbl, Clark, Hennigan, Noland, Millican, Driessche, Damoc, Guy, Osindero, Simonyan, Elsen, Rae, Vinyals, and Sifre]{hoffmann_training_2022}
Hoffmann, J., Borgeaud, S., Mensch, A., Buchatskaya, E., Cai, T., Rutherford, E., Casas, D. d.~L., Hendricks, L.~A., Welbl, J., Clark, A., Hennigan, T., Noland, E., Millican, K., Driessche, G. v.~d., Damoc, B., Guy, A., Osindero, S., Simonyan, K., Elsen, E., Rae, J.~W., Vinyals, O., and Sifre, L.
\newblock Training {Compute}-{Optimal} {Large} {Language} {Models}, March 2022.
\newblock URL \url{http://arxiv.org/abs/2203.15556}.
\newblock arXiv:2203.15556 [cs].

\bibitem[Huang et~al.(2019)Huang, Cheng, Bapna, Firat, Chen, Chen, Lee, Ngiam, Le, Wu, and Chen]{huang_gpipe_2019}
Huang, Y., Cheng, Y., Bapna, A., Firat, O., Chen, M.~X., Chen, D., Lee, H., Ngiam, J., Le, Q.~V., Wu, Y., and Chen, Z.
\newblock {GPipe}: efficient training of giant neural networks using pipeline parallelism.
\newblock In \emph{Proceedings of the 33rd {International} {Conference} on {Neural} {Information} {Processing} {Systems}}, number~10, pp.\  103--112. Curran Associates Inc., Red Hook, NY, USA, December 2019.

\bibitem[Jaech et~al.(2024)Jaech, Kalai, Lerer, Richardson, El-Kishky, Low, Helyar, Madry, Beutel, Carney, et~al.]{jaech2024openai}
Jaech, A., Kalai, A., Lerer, A., Richardson, A., El-Kishky, A., Low, A., Helyar, A., Madry, A., Beutel, A., Carney, A., et~al.
\newblock Openai o1 system card.
\newblock \emph{arXiv preprint arXiv:2412.16720}, 2024.

\bibitem[Jaghouar et~al.(2024)Jaghouar, Ong, Basra, Obeid, Straube, Keiblinger, Bakouch, Atkins, Panahi, Goddard, et~al.]{jaghouar2024intellect}
Jaghouar, S., Ong, J.~M., Basra, M., Obeid, F., Straube, J., Keiblinger, M., Bakouch, E., Atkins, L., Panahi, M., Goddard, C., et~al.
\newblock Intellect-1 technical report.
\newblock \emph{arXiv preprint arXiv:2412.01152}, 2024.

\bibitem[Korthikanti et~al.(2023)Korthikanti, Casper, Lym, McAfee, Andersch, Shoeybi, and Catanzaro]{korthikanti_reducing_2023}
Korthikanti, V.~A., Casper, J., Lym, S., McAfee, L., Andersch, M., Shoeybi, M., and Catanzaro, B.
\newblock Reducing activation recomputation in large transformer models.
\newblock \emph{Proceedings of Machine Learning and Systems}, 5:\penalty0 341--353, 2023.

\bibitem[Kudo \& Richardson(2018)Kudo and Richardson]{kudo_sentencepiece_2018}
Kudo, T. and Richardson, J.
\newblock {SentencePiece}: {A} simple and language independent subword tokenizer and detokenizer for {Neural} {Text} {Processing}.
\newblock In Blanco, E. and Lu, W. (eds.), \emph{Proceedings of the 2018 {Conference} on {Empirical} {Methods} in {Natural} {Language} {Processing}: {System} {Demonstrations}}, pp.\  66--71, Brussels, Belgium, November 2018. Association for Computational Linguistics.
\newblock \doi{10.18653/v1/D18-2012}.
\newblock URL \url{https://aclanthology.org/D18-2012/}.

\bibitem[Kulp et~al.(2024)Kulp, Gonzales, Smith, Heim, Puri, Vermeer, and Winkelman]{WR-A3056-1}
Kulp, G., Gonzales, D., Smith, E., Heim, L., Puri, P., Vermeer, M. J.~D., and Winkelman, Z.
\newblock \emph{Hardware-Enabled Governance Mechanisms: Developing Technical Solutions to Exempt Items Otherwise Classified Under Export Control Classification Numbers 3A090 and 4A090}.
\newblock RAND Corporation, Santa Monica, CA, 2024.
\newblock \doi{10.7249/WRA3056-1}.

\bibitem[Lambert et~al.(2024)Lambert, Morrison, Pyatkin, Huang, Ivison, Brahman, Miranda, Liu, Dziri, Lyu, et~al.]{lambert2024tulu}
Lambert, N., Morrison, J., Pyatkin, V., Huang, S., Ivison, H., Brahman, F., Miranda, L. J.~V., Liu, A., Dziri, N., Lyu, S., et~al.
\newblock T\"ulu 3: Pushing frontiers in open language model post-training.
\newblock \emph{arXiv preprint arXiv:2411.15124}, 2024.

\bibitem[Lamy-Poirier(2023)]{lamy-poirier_breadth-first_2023}
Lamy-Poirier, J.
\newblock Breadth-{First} {Pipeline} {Parallelism}.
\newblock \emph{Proceedings of Machine Learning and Systems}, 5:\penalty0 48--67, March 2023.
\newblock URL \url{{https://proceedings.mlsys.org/paper_files/paper/2023/hash/24e845415c1486dd2d582a9d639237f9-Abstract-mlsys2023.html}}.

\bibitem[Li et~al.(2024)Li, Fang, Smyrnis, Ivgi, Jordan, Gadre, Bansal, Guha, Keh, Arora, et~al.]{li2024datacomp}
Li, J., Fang, A., Smyrnis, G., Ivgi, M., Jordan, M., Gadre, S.~Y., Bansal, H., Guha, E., Keh, S.~S., Arora, K., et~al.
\newblock Datacomp-lm: In search of the next generation of training sets for language models.
\newblock \emph{Advances in Neural Information Processing Systems}, 37:\penalty0 14200--14282, 2024.

\bibitem[Liu et~al.(2024{\natexlab{a}})Liu, Feng, Wang, Wang, Liu, Zhao, Dengr, Ruan, Dai, Guo, et~al.]{deepseek-ai_deepseek-v2_2024}
Liu, A., Feng, B., Wang, B., Wang, B., Liu, B., Zhao, C., Dengr, C., Ruan, C., Dai, D., Guo, D., et~al.
\newblock {DeepSeek}-{V2}: {A} {Strong}, {Economical}, and {Efficient} {Mixture}-of-{Experts} {Language} {Model}, 2024{\natexlab{a}}.

\bibitem[Liu et~al.(2024{\natexlab{b}})Liu, Feng, Xue, Wang, Wu, Lu, Zhao, Deng, Zhang, Ruan, et~al.]{deepseek-ai_deepseek-v3_2025}
Liu, A., Feng, B., Xue, B., Wang, B., Wu, B., Lu, C., Zhao, C., Deng, C., Zhang, C., Ruan, C., et~al.
\newblock {DeepSeek-V3 Technical Report}, 2024{\natexlab{b}}.

\bibitem[Liu et~al.(2023)Liu, Zaharia, and Abbeel]{liu_ringattention_2023}
Liu, H., Zaharia, M., and Abbeel, P.
\newblock {RingAttention} with {Blockwise} {Transformers} for {Near}-{Infinite} {Context}.
\newblock October 2023.
\newblock URL \url{https://openreview.net/forum?id=WsRHpHH4s0}.

\bibitem[Loshchilov \& Hutter(2018)Loshchilov and Hutter]{loshchilov_decoupled_2018}
Loshchilov, I. and Hutter, F.
\newblock Decoupled {Weight} {Decay} {Regularization}.
\newblock September 2018.
\newblock URL \url{https://openreview.net/forum?id=Bkg6RiCqY7}.

\bibitem[Miotti et~al.(2024)Miotti, Bilge, Kasten, and Newport]{miotti_narrow_2024}
Miotti, A., Bilge, T., Kasten, D., and Newport, J.
\newblock A {Narrow} {Path}, December 2024.
\newblock URL \url{https://www.narrowpath.co/}.

\bibitem[Narayanan et~al.(2021)Narayanan, Shoeybi, Casper, LeGresley, Patwary, Korthikanti, Vainbrand, Kashinkunti, Bernauer, Catanzaro, Phanishayee, and Zaharia]{narayanan_efficient_2021}
Narayanan, D., Shoeybi, M., Casper, J., LeGresley, P., Patwary, M., Korthikanti, V., Vainbrand, D., Kashinkunti, P., Bernauer, J., Catanzaro, B., Phanishayee, A., and Zaharia, M.
\newblock Efficient large-scale language model training on {GPU} clusters using megatron-{LM}.
\newblock In \emph{Proceedings of the {International} {Conference} for {High} {Performance} {Computing}, {Networking}, {Storage} and {Analysis}}, pp.\  1--15, St. Louis Missouri, November 2021. ACM.
\newblock ISBN 978-1-4503-8442-1.
\newblock \doi{10.1145/3458817.3476209}.
\newblock URL \url{https://dl.acm.org/doi/10.1145/3458817.3476209}.

\bibitem[Noune et~al.(2022)Noune, Jones, Justus, Masters, and Luschi]{noune_8-bit_2022}
Noune, B., Jones, P., Justus, D., Masters, D., and Luschi, C.
\newblock 8-bit {Numerical} {Formats} for {Deep} {Neural} {Networks}, June 2022.
\newblock URL \url{http://arxiv.org/abs/2206.02915}.
\newblock arXiv:2206.02915 [cs].

\bibitem[{OLMo Team} et~al.(2024){OLMo Team}, Walsh, Soldaini, Groeneveld, Lo, Arora, Bhagia, Gu, Huang, Jordan, et~al.]{olmo20242olmo}
{OLMo Team}, Walsh, P., Soldaini, L., Groeneveld, D., Lo, K., Arora, S., Bhagia, A., Gu, Y., Huang, S., Jordan, M., et~al.
\newblock 2 olmo 2 furious.
\newblock \emph{arXiv preprint arXiv:2501.00656}, 2024.

\bibitem[Ouyang et~al.(2022)Ouyang, Wu, Jiang, Almeida, Wainwright, Mishkin, Zhang, Agarwal, Slama, Ray, et~al.]{ouyang2022training}
Ouyang, L., Wu, J., Jiang, X., Almeida, D., Wainwright, C., Mishkin, P., Zhang, C., Agarwal, S., Slama, K., Ray, A., et~al.
\newblock Training language models to follow instructions with human feedback.
\newblock \emph{Advances in neural information processing systems}, 35:\penalty0 27730--27744, 2022.

\bibitem[Peng et~al.(2023)Peng, Wu, Wei, Zhao, Yang, Liu, Xiong, Yang, Ni, Hu, Li, Zhang, Li, Ning, Wang, Zhang, Liu, Chau, Hu, and Cheng]{peng_fp8-lm_2023}
Peng, H., Wu, K., Wei, Y., Zhao, G., Yang, Y., Liu, Z., Xiong, Y., Yang, Z., Ni, B., Hu, J., Li, R., Zhang, M., Li, C., Ning, J., Wang, R., Zhang, Z., Liu, S., Chau, J., Hu, H., and Cheng, P.
\newblock {FP8}-{LM}: {Training} {FP8} {Large} {Language} {Models}, December 2023.
\newblock URL \url{http://arxiv.org/abs/2310.18313}.
\newblock arXiv:2310.18313 [cs].

\bibitem[Petrie et~al.(2024)Petrie, Aarne, Ammann, and Dalrymple]{petrie2024interim}
Petrie, J., Aarne, O., Ammann, N., and Dalrymple, D.
\newblock Interim report: Mechanisms for flexible hardware-enabled guarantees.
\newblock Technical report, 8 2024.

\bibitem[Qi et~al.(2023)Qi, Wan, Huang, and Lin]{qi_zero_2023}
Qi, P., Wan, X., Huang, G., and Lin, M.
\newblock Zero {Bubble} ({Almost}) {Pipeline} {Parallelism}.
\newblock October 2023.
\newblock URL \url{https://openreview.net/forum?id=tuzTN0eIO5}.

\bibitem[Rabe \& Staats(2021)Rabe and Staats]{rabe_self-attention_2022}
Rabe, M.~N. and Staats, C.
\newblock Self-attention does not need $ o (n^2) $ memory, 2021.

\bibitem[Rajbhandari et~al.(2020)Rajbhandari, Rasley, Ruwase, and He]{rajbhandari_zero_2020}
Rajbhandari, S., Rasley, J., Ruwase, O., and He, Y.
\newblock {ZeRO}: memory optimizations toward training trillion parameter models.
\newblock In \emph{Proceedings of the {International} {Conference} for {High} {Performance} {Computing}, {Networking}, {Storage} and {Analysis}}, {SC} '20, pp.\  1--16, Atlanta, Georgia, November 2020. IEEE Press.
\newblock ISBN 978-1-72819-998-6.

\bibitem[Ren et~al.(2021)Ren, Rajbhandari, Aminabadi, Ruwase, Yang, Zhang, Li, and He]{ren_zero-offload_2021}
Ren, J., Rajbhandari, S., Aminabadi, R.~Y., Ruwase, O., Yang, S., Zhang, M., Li, D., and He, Y.
\newblock {ZeRO}-{Offload}: {Democratizing} {Billion}-{Scale} {Model} {Training}.
\newblock January 2021.
\newblock URL \url{https://openreview.net/forum?id=qXFQtGMHRa}.

\bibitem[Sastry et~al.(2024)Sastry, Heim, Belfield, Anderljung, Brundage, Hazell, O'Keefe, Hadfield, Ngo, Pilz, et~al.]{sastry2024computing}
Sastry, G., Heim, L., Belfield, H., Anderljung, M., Brundage, M., Hazell, J., O'Keefe, C., Hadfield, G.~K., Ngo, R., Pilz, K., et~al.
\newblock Computing power and the governance of artificial intelligence.
\newblock \emph{arXiv preprint arXiv:2402.08797}, 2024.

\bibitem[Scher \& Thiergart(2024)Scher and Thiergart]{scher_mechanisms_2024}
Scher, A. and Thiergart, L.
\newblock Mechanisms to {Verify} {International} {Agreements} {About} {AI} {Development}, November 2024.
\newblock URL \url{https://techgov.intelligence.org/research/mechanisms-to-verify-international-agreements-about-ai-development}.

\bibitem[Sennrich et~al.(2016)Sennrich, Haddow, and Birch]{sennrich_neural_2016}
Sennrich, R., Haddow, B., and Birch, A.
\newblock Neural {Machine} {Translation} of {Rare} {Words} with {Subword} {Units}.
\newblock In Erk, K. and Smith, N.~A. (eds.), \emph{Proceedings of the 54th {Annual} {Meeting} of the {Association} for {Computational} {Linguistics} ({Volume} 1: {Long} {Papers})}, pp.\  1715--1725, Berlin, Germany, August 2016. Association for Computational Linguistics.
\newblock \doi{10.18653/v1/P16-1162}.
\newblock URL \url{https://aclanthology.org/P16-1162/}.

\bibitem[Shazeer(2020)]{shazeer_glu_2020}
Shazeer, N.
\newblock {GLU} {Variants} {Improve} {Transformer}, February 2020.
\newblock URL \url{http://arxiv.org/abs/2002.05202}.
\newblock arXiv:2002.05202 [cs].

\bibitem[Shoeybi et~al.(2020)Shoeybi, Patwary, Puri, LeGresley, Casper, and Catanzaro]{shoeybi_megatron-lm_2020}
Shoeybi, M., Patwary, M., Puri, R., LeGresley, P., Casper, J., and Catanzaro, B.
\newblock Megatron-{LM}: {Training} {Multi}-{Billion} {Parameter} {Language} {Models} {Using} {Model} {Parallelism}, March 2020.
\newblock URL \url{http://arxiv.org/abs/1909.08053}.
\newblock arXiv:1909.08053 [cs].

\bibitem[Stiennon et~al.(2020)Stiennon, Ouyang, Wu, Ziegler, Lowe, Voss, Radford, Amodei, and Christiano]{stiennon2020learning}
Stiennon, N., Ouyang, L., Wu, J., Ziegler, D., Lowe, R., Voss, C., Radford, A., Amodei, D., and Christiano, P.~F.
\newblock Learning to summarize with human feedback.
\newblock \emph{Advances in neural information processing systems}, 33:\penalty0 3008--3021, 2020.

\bibitem[Su et~al.(2024)Su, Ahmed, Lu, Pan, Bo, and Liu]{su_roformer_2024}
Su, J., Ahmed, M., Lu, Y., Pan, S., Bo, W., and Liu, Y.
\newblock {RoFormer}: {Enhanced} transformer with {Rotary} {Position} {Embedding}.
\newblock \emph{Neurocomputing}, 568:\penalty0 127063, February 2024.
\newblock ISSN 0925-2312.
\newblock \doi{10.1016/j.neucom.2023.127063}.
\newblock URL \url{https://www.sciencedirect.com/science/article/pii/S0925231223011864}.

\bibitem[Touvron et~al.(2023{\natexlab{a}})Touvron, Lavril, Izacard, Martinet, Lachaux, Lacroix, Rozi{\`e}re, Goyal, Hambro, Azhar, et~al.]{touvron2023llama1}
Touvron, H., Lavril, T., Izacard, G., Martinet, X., Lachaux, M.-A., Lacroix, T., Rozi{\`e}re, B., Goyal, N., Hambro, E., Azhar, F., et~al.
\newblock Llama: Open and efficient foundation language models.
\newblock \emph{arXiv preprint arXiv:2302.13971}, 2023{\natexlab{a}}.

\bibitem[Touvron et~al.(2023{\natexlab{b}})Touvron, Martin, Stone, Albert, Almahairi, Babaei, Bashlykov, Batra, Bhargava, Bhosale, et~al.]{touvron2023llama2}
Touvron, H., Martin, L., Stone, K., Albert, P., Almahairi, A., Babaei, Y., Bashlykov, N., Batra, S., Bhargava, P., Bhosale, S., et~al.
\newblock Llama 2: Open foundation and fine-tuned chat models.
\newblock \emph{arXiv preprint arXiv:2307.09288}, 2023{\natexlab{b}}.

\bibitem[Vaswani et~al.(2017)Vaswani, Shazeer, Parmar, Uszkoreit, Jones, Gomez, Kaiser, and Polosukhin]{vaswani_attention_2017}
Vaswani, A., Shazeer, N., Parmar, N., Uszkoreit, J., Jones, L., Gomez, A.~N., Kaiser, {\L}., and Polosukhin, I.
\newblock Attention is all you need.
\newblock volume~30, 2017.

\bibitem[Wang et~al.(2024)Wang, Gao, Zhao, Sun, and Dai]{wang_auxiliary-loss-free_2024}
Wang, L., Gao, H., Zhao, C., Sun, X., and Dai, D.
\newblock Auxiliary-{Loss}-{Free} {Load} {Balancing} {Strategy} for {Mixture}-of-{Experts}.
\newblock October 2024.
\newblock URL \url{https://openreview.net/forum?id=y1iU5czYpE}.

\bibitem[Wenzek et~al.(2019)Wenzek, Lachaux, Conneau, Chaudhary, Guzm{\'a}n, Joulin, and Grave]{wenzek2019ccnet}
Wenzek, G., Lachaux, M.-A., Conneau, A., Chaudhary, V., Guzm{\'a}n, F., Joulin, A., and Grave, E.
\newblock Ccnet: Extracting high quality monolingual datasets from web crawl data.
\newblock \emph{arXiv preprint arXiv:1911.00359}, 2019.

\bibitem[Xiong et~al.(2024)Xiong, Liu, Molybog, Zhang, Bhargava, Hou, Martin, Rungta, Sankararaman, Oguz, Khabsa, Fang, Mehdad, Narang, Malik, Fan, Bhosale, Edunov, Lewis, Wang, and Ma]{xiong_effective_2024}
Xiong, W., Liu, J., Molybog, I., Zhang, H., Bhargava, P., Hou, R., Martin, L., Rungta, R., Sankararaman, K.~A., Oguz, B., Khabsa, M., Fang, H., Mehdad, Y., Narang, S., Malik, K., Fan, A., Bhosale, S., Edunov, S., Lewis, M., Wang, S., and Ma, H.
\newblock Effective {Long}-{Context} {Scaling} of {Foundation} {Models}.
\newblock In Duh, K., Gomez, H., and Bethard, S. (eds.), \emph{Proceedings of the 2024 {Conference} of the {North} {American} {Chapter} of the {Association} for {Computational} {Linguistics}: {Human} {Language} {Technologies} ({Volume} 1: {Long} {Papers})}, pp.\  4643--4663, Mexico City, Mexico, June 2024. Association for Computational Linguistics.
\newblock \doi{10.18653/v1/2024.naacl-long.260}.
\newblock URL \url{https://aclanthology.org/2024.naacl-long.260/}.

\bibitem[Zhang \& Sennrich(2019)Zhang and Sennrich]{zhang_root_2019}
Zhang, B. and Sennrich, R.
\newblock Root mean square layer normalization.
\newblock In \emph{Proceedings of the 33rd {International} {Conference} on {Neural} {Information} {Processing} {Systems}}, number 1110, pp.\  12381--12392. Curran Associates Inc., Red Hook, NY, USA, December 2019.

\bibitem[Zhao et~al.(2023)Zhao, Gu, Varma, Luo, Huang, Xu, Wright, Shojanazeri, Ott, Shleifer, Desmaison, Balioglu, Damania, Nguyen, Chauhan, Hao, Mathews, and Li]{zhao_pytorch_2023}
Zhao, Y., Gu, A., Varma, R., Luo, L., Huang, C.-C., Xu, M., Wright, L., Shojanazeri, H., Ott, M., Shleifer, S., Desmaison, A., Balioglu, C., Damania, P., Nguyen, B., Chauhan, G., Hao, Y., Mathews, A., and Li, S.
\newblock {PyTorch} {FSDP}: {Experiences} on {Scaling} {Fully} {Sharded} {Data} {Parallel}, September 2023.
\newblock URL \url{http://arxiv.org/abs/2304.11277}.
\newblock arXiv:2304.11277 [cs].

\end{thebibliography}
\bibliographystyle{icml2025}

\newpage
\appendix
\onecolumn

\renewcommand{\arraystretch}{1.3} 

\begin{landscape}
\section{Table of algorithmic innovations}
\label{app:table_of_innovations}
\begin{longtable}{@{} p{4cm} l l r S[table-auto-round] l c l l @{}} 

\caption{Algorithmic Innovations in Large Language Models. \llamathree{} shows innovations used in Llama 3 developed by Meta, \deepseekvthree{} shows innovations used in DeepSeek-V3.  The \emph{Math equiv} column shows innovations which are more computationally efficient ways of performing mathematically equivalent operations. } 
\label{tab:algo_innovations}\\

\toprule
\bfseries Innovation & \bfseries Date & \bfseries Models & {\bfseries FLOP}  & {\bfseries TFLOP/s} & \bfseries Category & \bfseries Math equiv & \bfseries Sector & \bfseries Country \\
\midrule
\endfirsthead 

\caption[]{Algorithmic Innovations in Large Language Models -- Continued}\\
\toprule
\bfseries Innovation & \bfseries Date & \bfseries Models & {\bfseries FLOP}  & {\bfseries TFLOP/s} & \bfseries Category & \bfseries Math equiv & \bfseries Sector & \bfseries Country \\
\midrule
\endhead 

\multicolumn{9}{r}{\textit{Continued on next page}} \\
\bottomrule 
\endfoot 

\bottomrule 
\endlastfoot 

Transformer architecture \citep{vaswani_attention_2017} & 6/12/2017 & \llamathree, \deepseekvthree & $4.00 \times 10^{19}$ & 84.8 & Architecture &  & Industry & USA \\ \hline
RMSNorm \citep{zhang_root_2019} & 10/16/2019 & \llamathree & $1.37 \times 10^{20}$ & 149.42 & Architecture &  & Academic & UK \\ \hline
SwiGLU \citep{shazeer_glu_2020} & 2/12/2020 & \llamathree & $5.56 \times 10^{20}$ & 1440 & Architecture &  & Industry & USA \\ \hline
Rotary Position Embedding \citep{su_roformer_2024} & 2/3/2021 & \llamathree, \deepseekvthree & $9.62 \times 10^{19}$ & 1000 & Architecture &  & Industry & China \\ \hline
Grouped Query Attention \citep{ainslie_gqa_2023} & 5/22/2023 & \llamathree & $1.91 \times 10^{21}$ & 2200 & Architecture &  & Industry & USA \\ \hline
Increasing RoPE Base Frequency \citep{xiong_effective_2024} & 9/27/2023 & \llamathree & $1.01 \times 10^{22}$ & \nodata & Architecture &  & Industry & USA \\ \hline
DeepSeekMoE \citep{dai2024deepseekmoe} & 1/11/2024 & \deepseekvthree & $1.19 \times 10^{23}$ & \nodata & Architecture &  & Industry & China \\ \hline
Multi-head Latent Attention (MLA) \citep{deepseek-ai_deepseek-v2_2024} & 5/7/2024 & \deepseekvthree & $1.56 \times 10^{23}$ & \nodata & Architecture &  & Industry & China \\ \hline
Byte-pair encoding \citep{sennrich_neural_2016} & 8/31/2015 & \llamathree, \deepseekvthree & $9.00 \times 10^{18}$ & 22.58 & Data \& Tokenization &  & Academic & UK \\ \hline
SentencePiece \citep{kudo_sentencepiece_2018} & 8/19/2018 & \llamathree & \negligible & \nodata & Data \& Tokenization &  & Industry & USA \\ \hline
Line-level de-duplication \citep{wenzek2019ccnet} & 11/1/2019 & \llamathree & $2.76 \times 10^{21}$ & 2000 & Data \& Tokenization &  & Industry & USA \\ \hline
Annealing on high quality data \citep{li2024datacomp} & 5/17/2024 & \llamathree & $1.25 \times 10^{23}$ & \nodata & Data \& Tokenization &  & Academic & USA \\ \hline
Warp specialization \citep{bauer_singe_2014} & 2/6/2014 & \deepseekvthree & \negligible & 4.528 & Efficiencies & \mathequivcheck & Academic & USA \\ \hline
Efficient implementation of the causal multi-head attention \citep{rabe_self-attention_2022} & 12/10/2021 & \llamathree & $1.95 \times 10^{18}$ & 123 & Efficiencies & \mathequivcheck & Industry & USA \\ \hline
Sequence parallelism and selective activation recomputation \citep{korthikanti_reducing_2023} & 5/10/2022 & \llamathree & \negligible & 349440 & Efficiencies & \mathequivcheck & Industry & USA \\ \hline
FlashAttention \citep{dao_flashattention_2022} & 5/27/2022 & \llamathree & $6.22 \times 10^{20}$ & 2539.721 & Efficiencies & \mathequivcheck & Academic & USA \\ \hline
8-bit Numerical Formats for Deep Neural Networks \citep{noune_8-bit_2022} & 6/6/2022 & \deepseekvthree & $5.51 \times 10^{21}$ & \nodata & Efficiencies &  & Industry & UK \\ \hline
LLM.int8() \citep{dettmers_llmint8_2022} & 8/15/2022 & \deepseekvthree & $1.39 \times 10^{21}$ & 1248 & Efficiencies &  & Industry & USA \\ \hline
FP8-LM \citep{peng_fp8-lm_2023} & 10/27/2023 & \deepseekvthree & $1.08 \times 10^{23}$ & 15824 & Efficiencies &  & Industry & USA \\ \hline
Chinchilla scaling laws \citep{hoffmann_training_2022} & 3/29/2022 & \llamathree & $1.05 \times 10^{23}$ & \nodata & Scaling laws &  & Industry & UK, USA \\ \hline
LLaMA 3 scaling laws \citep{grattafiori_llama_2024} & 7/23/2024 & \llamathree & $6.60 \times 10^{22}$ & \nodata & Scaling laws &  & Industry & USA \\ \hline
GPipe: Pipeline Parallelism \citep{huang_gpipe_2019} & 11/16/2018 & \llamathree & $1.85 \times 10^{21}$ & 16188.8 & Scaling training & \mathequivcheck & Industry & USA \\ \hline
Tensor parallelism \citep{shoeybi_megatron-lm_2020} & 9/17/2019 & \llamathree & $6.98 \times 10^{22}$ & 64000 & Scaling training & \mathequivcheck & Industry & USA \\ \hline
Zero Redundancy Optimizer (ZeRO) \citep{rajbhandari_zero_2020} & 10/4/2019 & \llamathree, \deepseekvthree & \negligible & 50000 & Scaling training & \mathequivcheck & Industry & USA \\ \hline
ZeRO-Offload \citep{ren_zero-offload_2021} & 1/18/2021 & \llamathree & \negligible & 16000 & Scaling training & \mathequivcheck & Industry & USA \\ \hline
PTD-P \citep{narayanan_efficient_2021} & 4/9/2021 & \llamathree & \negligible & 479232 & Scaling training & \mathequivcheck & Industry & USA \\ \hline
Breadth-First Pipeline Parallelism \citep{lamy-poirier_breadth-first_2023} & 11/11/2022 & \llamathree & \negligible & 8000 & Scaling training & \mathequivcheck & Industry & Canada \\ \hline
Fully sharded data parallelism \citep{zhao_pytorch_2023} & 4/21/2023 & \llamathree & \negligible & 79872 & Scaling training & \mathequivcheck & Industry & USA \\ \hline
Context Parallelism \citep{liu_ringattention_2023} & 10/3/2023 & \llamathree & $9.65 \times 10^{20}$ & 400000 & Scaling training & \mathequivcheck & Academic & USA \\ \hline
Pipeline Parallelism (PP) \citep{qi_zero_2023} & 11/30/2023 & \deepseekvthree & \negligible & 4992 & Scaling training & \mathequivcheck & Industry & Singapore \\ \hline
AdamW \citep{loshchilov_decoupled_2018} & 11/14/2017 & \llamathree, \deepseekvthree & $1.76 \times 10^{19}$ & 329.1 & Training &  & Academic & Germany \\ \hline
Fill-in-Middle \citep{bavarian_efficient_2022} & 7/28/2022 & \deepseekvthree & $6.55 \times 10^{22}$ & \nodata & Training &  & Industry & USA \\ \hline
Multi-token Prediction \citep{gloeckle_better_2024} & 4/30/2024 & \deepseekvthree & $3.85 \times 10^{23}$ & \nodata & Training &  & Industry & USA, France \\ \hline
Fill-in-Middle DeepSeek validation \citep{deepseek-ai_deepseek-coder-v2_2024} & 6/17/2024 & \deepseekvthree & $8.64 \times 10^{22}$ & \nodata & Training &  & Industry & China \\ \hline
Auxiliary-loss-free load balancing \citep{wang_auxiliary-loss-free_2024} & 8/28/2024 & \deepseekvthree & $2.76 \times 10^{21}$ & \nodata & Training &  & Industry & China \\ \hline
Multi-Token Prediction (MTP) \citep{deepseek-ai_deepseek-v3_2025} & 12/27/2024 & \deepseekvthree & $1.74 \times 10^{23}$ & \nodata & Training &  & Industry & China \\

\end{longtable}

\end{landscape}


\end{document}